\documentclass[10pt,letterpaper]{article}

\usepackage{ccn}
\usepackage{pslatex}
\usepackage{apacite}
\usepackage{graphicx}
\usepackage[font=footnotesize]{subfig}

\title{The effect of task and training on intermediate representations in convolutional neural networks revealed with modified RV similarity analysis}
 
\author{{\large \bf Jessica A.F. Thompson (j.thompson@umontreal.ca)} \\
  BRAMS, CRBLM, Mila, Universit\'{e} de Montr\'{e}al\\
Montreal, Quebec, Canada
  \AND {\large \bf Yoshua Bengio (yoshua.bengio@mila.quebec)} \\
  Mila, Universit\'{e} de Montr\'{e}al\\
  Montreal, Quebec, Canada
\AND {\large \bf Marc Sch\"onwiesner (marcs@rz.uni-leipzig.de)} \\
  Department of Biology, University of Leipzig\\
  Leipzig, Germany}

\begin{document}

\maketitle

\section{Abstract}
{
\bf Centered Kernel Alignment (CKA) was recently proposed as a similarity metric for comparing activation patterns in deep networks. Here we experiment with the modified RV-coefficient (RV2), which has similar properties to CKA while being less sensitive to dataset size. We compare the representations of networks that received varying amounts of training on different layers: a standard trained network (all parameters updated at every step), a freeze-trained network (layers gradually frozen during training), random networks (only some layers trained), and a completely untrained network. We found that RV2 was able to recover expected similarity patterns and provide interpretable similarity matrices that suggested hypotheses about how representations are affected by different training recipes. We propose that the superior performance achieved by freeze training can be attributed to representational differences in the penultimate layer. Comparisons to random networks suggest that the inputs and targets serve as anchors on the representations in the lowest and highest layers.

}
\begin{quote}
\small
\textbf{Keywords:} 
similarity analysis, random features, CNNs, freeze training, RV coefficient
\end{quote}

\section{Introduction}
The study of artificial and biological neural networks often requires quantification of the similarity of activation patterns between two networks. Common approaches to this problem are variants of canonical correlation analysis (CCA) \cite{Hotelling1936RelationsVariates}. For example, Singular Vector CCA and Projection-Weighted CCA have recently been used to uncover insights about training dynamics and generalization in deep networks \cite{Raghu2017, Morcos2018}. Regularized CCA is often used in neuroscience to find relationships between neural and behavioural or clinical variables \cite{Bilenko2016Pyrcca:Neuroimaging}. However, these variants of CCA can require large amounts of data and so are often impractical for analyzing neural activations where the number of observations may be small and the dimensionality may be large.

When comparing two sets of variables $\mathbf{X}$ and $\mathbf{Y}$, CCA will find the linear combinations of $\mathbf{X}$ and $\mathbf{Y}$ which maximizes their correlation. This means that CCA is invariant to any invertible linear transformation. There are several reasons why one might want a similarity metric with different invariance properties. For example, in a deep network, it is not just the linear information content of a representation that is meaningful but also the specific configuration of that information. For example, the insertion of an invertible linear transformation between two layers of a deep network can alter the network's behaviour (e.g. in batch normalization). Therefore, when comparing representations in deep neural networks, one may wish to use a similarity metric that is not invariant to invertible linear transformation so as to be sensitive to meaningful differences between representations \cite{Kornblith2019SimilarityRevisited, Thompson2016HowProcessing}.

\citeA{Kornblith2019SimilarityRevisited} propose the use of Centered Kernel Alignment (CKA) based on the fact that CKA is only invariant to orthogonal transformations and isomorphic scaling (not arbitrary linear invertible transformations) and that it demonstrates intuitive notions of similarity, namely that corresponding layers are most similar to themselves in networks of identical architecture trained from different random initializations. 
They state that CKA with a linear kernel is equivalent to the RV coefficient. 
The RV coefficient is a matrix correlation method for comparing paired observations $\mathbf{X}$ and $\mathbf{Y}$ with different numbers of columns \cite{Robert1976ACoefficient}. 
\begin{equation}
    RV(\mathbf{X}, \mathbf{Y}) = \frac{tr(\mathbf{XX}^\prime\mathbf{YY}^\prime)}{\sqrt{tr[(\mathbf{XX}^\prime)^2]tr[(\mathbf{YY}^\prime)^2]}}
\end{equation}

The RV coefficient is still sensitive to dataset size. When the number of observations is too small relative to the number of dimensions, the RV coefficient will tend to $1$, even for random, unrelated matrices. The modified RV coefficient (RV2) addresses this problem by ignoring the diagonal elements of $\mathbf{XX^\prime}$ and $\mathbf{YY^\prime}$, which pushes the numerator to zero when $\mathbf{X}$ and $\mathbf{Y}$ are random matrices, even for small sample sizes \cite{Smilde2009MatrixRV-coefficient}.
\begin{equation}
    RV_2(\mathbf{X}, \mathbf{Y}) = \frac{Vec(\mathbf{\widetilde{XX^\prime}})^\prime Vec(\mathbf{\widetilde{YY^\prime}})}{\sqrt{Vec(\mathbf{\widetilde{XX^\prime}})^\prime Vec(\mathbf{\widetilde{XX^\prime}}) \times Vec(\mathbf{\widetilde{YY^\prime}})^\prime Vec(\mathbf{\widetilde{YY^\prime}})}}
\end{equation}
Where $\mathbf{\widetilde{XX^\prime}} = \mathbf{XX^\prime} - diag(\mathbf{XX^\prime})$ and similarly for $\mathbf{\widetilde{YY^\prime}}$. Thus RV2 provides a similarity metric with the same invariance properties as CKA while being less sensitive to dataset size, making it a good candidate for comparing neural activities of large artificial and biological neural networks. 

Here we explore the use of RV2 to characterize intermediate representations of simple convolutional neural networks. Our main contributions are (a) extending \citeA{Kornblith2019SimilarityRevisited}'s validation of CKA-flavored similarity metrics by using RV2 to recover expected similarity patterns in simple networks, and (b) showing that RV2 can generate interpretable patterns that can suggest hypotheses about the nature of intermediate representations in deep neural networks. 

\section{Experiments}
Trained networks in the following analyses were previously reported in \citeA{Thompson2019HowLanguages}. All networks were of identical architecture consisting of nine convolutional layers and three fully connected layers. Networks were trained to recognize context-dependent English or Dutch phones for 100 epochs (except for the untrained network). Networks differed in the training that they received. The standard networks were randomly initialized and all parameters were updated on every mini-batch. The untrained network was randomly initialized and never trained. The procedures for the freeze-trained and random networks are described below. Please refer to the original text for details about the datasets, architecture and training.

Activations to one hour of English speech from 60 speakers (1-minute each) were measured from all networks. We used the hoggorm python package to calculate RV2 for all pairs of layers. To make the experiments feasible, we performed average-pooling on all feature maps and downsampled the resulting activation vectors by a factor of 40, leading to activation vectors of length 23,582 per `unit'. 

\subsection{Untrained vs Trained}
\begin{figure}[h!]
\centering
\includegraphics[width=\columnwidth]{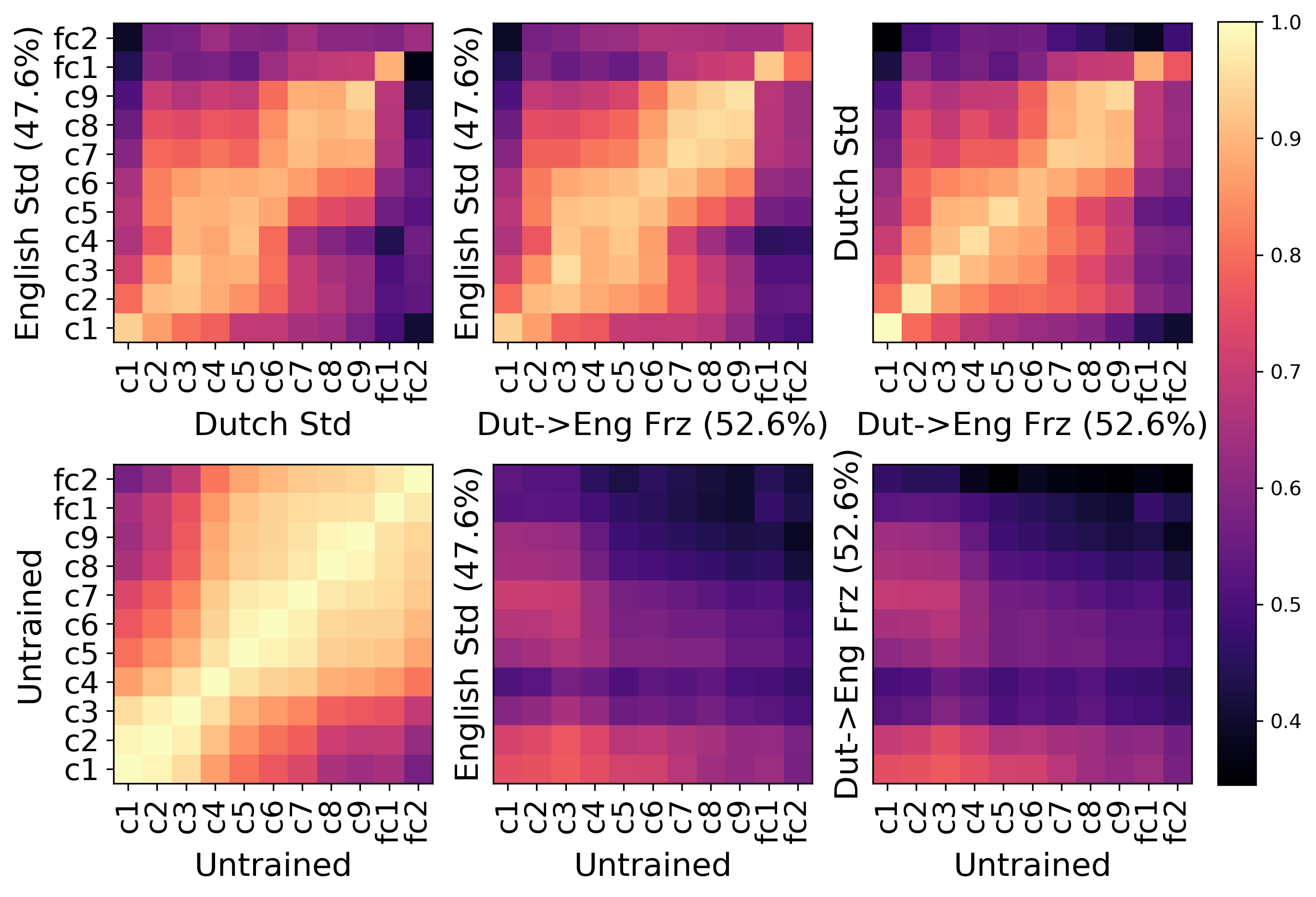}
\caption{\small \textbf{Top row} Similarity between English and Dutch standard networks and the Dutch-to-English transfer freeze-trained network. The largest differences are in fc2. Lower layers in the transfer freeze-trained network are most similar to their corresponding layer in the Dutch standard model.  \textbf{Bottom row} Network self-similarity at initialization (left) and the similarity between untrained and trained networks, either standard net (middle) or the transfer freeze-trained net (right). The parenthetical percentages indicates the top-1 accuracy.}
\label{fig:untrained-vs-trained}
\end{figure}

We replicated Figure~F.4 from \citeA{Kornblith2019SimilarityRevisited} to verify that a slightly different metric, RV2, applied to activations from a different model trained on a different task generates similar patterns of similarity between trained and untrained networks. Figure~\ref{fig:untrained-vs-trained} (bottom row) shows the self-similarity of an untrained network and the similarity between the untrained network and two different trained networks: standard training and transfer freeze-training (described in the next section). We observe approximately the same patterns as are reported in \citeA{Kornblith2019SimilarityRevisited}.
\subsection{Freeze Training}
It has been suggested that convolutional neural networks converge `bottom-up', with early layers converging to their final form earlier in training \cite{Raghu2017, Alain2016}. Based on this observation, \citeA{Raghu2017} proposed \textit{freeze training}. During freeze training, at regular intervals, the parameters of an additional layer are frozen (i.e. removed from the set of trainable variables). Layers are frozen in order by depth such that, by the end of training, only the final layer is being updated. The freeze-trained transfer networks from \citeA{Thompson2019HowLanguages}, which were initialized with parameters from a network previously trained on one language and then freeze-trained on another, outperformed all other freeze-trained networks (no transfer) and other transfer networks (no freeze training). Here, we compare the activations of the English standard, Dutch standard and Dutch-to-English freeze-trained networks from \citeA{Thompson2019HowLanguages}. We predict with high confidence that the early layers of the Dutch-to-English freeze-trained network will be more similar to the Dutch than the English standard model since they were initialized with the parameters from the Dutch standard network and received relatively little training afterwards. This provides a good test of whether RV2 is able to recover this expected pattern. Additionally, we were interested to see if the superior performance of the transfer freeze-trained network could be attributed to any representational differences between the compared networks.

For all comparisons between the standard and transfer freeze-trained networks (Figure~\ref{fig:untrained-vs-trained}, top row), the highest similarity values were near the diagonal.
This pattern provides further validation that, like CKA, RV2 finds the most similar layer in one network to be near the corresponding layer in another network of identical architecture. 
As predicted, early layers in the Dutch-to-English freeze-trained network were most similar to the corresponding layer in the Dutch standard model and less similar to the English standard model. Near corresponding layers in the English and Dutch standard models were considerably similar to one another, despite being trained on different languages. The largest differences in all comparisons occured in layer fc2. Thus, the superior performance of the transfer freeze-trained network may be primarily attributable to differences in representation at fc2.
%

\subsection{Random Features}
\begin{figure}[hb]
    \centering
    \includegraphics[width={\columnwidth}]{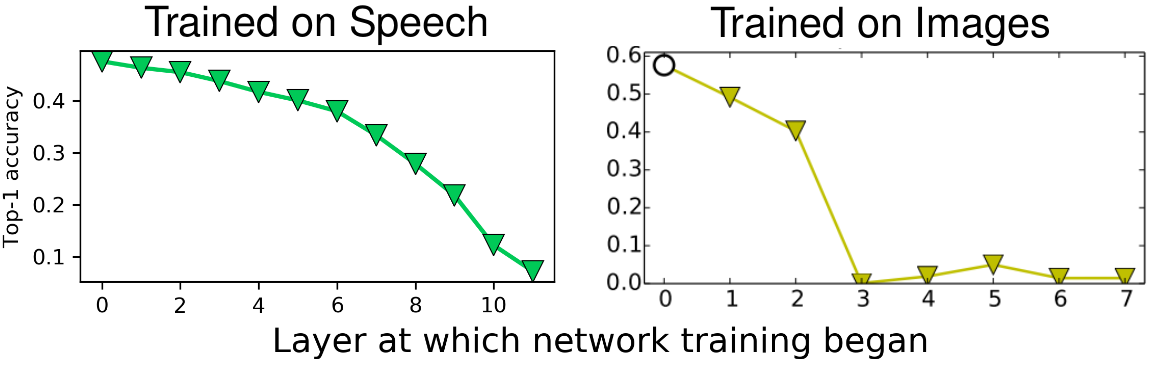}
    \caption{\small Performance of random networks as reported in \citeA{Thompson2019HowLanguages} (left) and \citeA{Yosinski2014} (right).}
    \label{fig:rand}
\end{figure}

\citeA{Yosinski2014} investigated the effect on performance of leaving progressively more layers untrained in convolutional neural networks trained to recognize objects in images. 
Performance dropped sharply to zero when the first three layers were left at their random initialization and only subsequent layers were trained. \citeA{Thompson2019HowLanguages} replicated this experiment with networks trained on speech and found a different pattern (see Figure~\ref{fig:rand}).
Performance gradually declined as more layers were left untrained, only reaching near-zero performance when all but the last layer were left untrained \cite{Thompson2019HowLanguages}. 

Random features have a long history of success in kernel machines \cite{Rahimi2007}. 
However, the effect of several consecutive random layers is less well understood. In particular, how do intermediate representations reconfigure as more layers are left untrained?

\begin{figure}[h]
\centering
\includegraphics[width=\columnwidth]{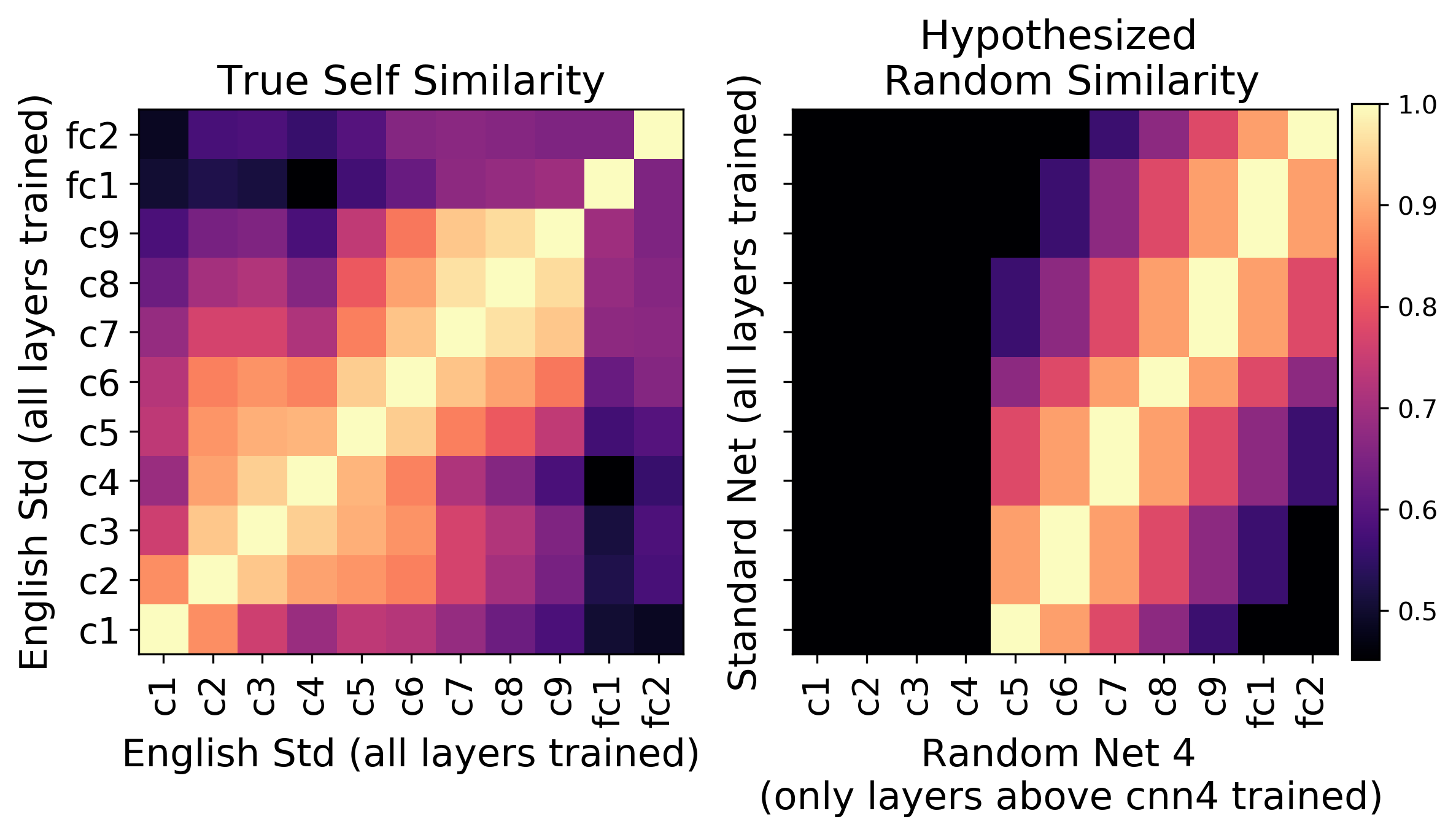}
\caption{\small \textbf{Left} Self-similarity of the English standard network. \textbf{Right} Idealized diagram of the hypothesis that the representational transformations of the standard network will be evenly distributed across the trained layers of a random net. } 
\label{fig:hypothesis}
\end{figure}

We presume that the effect of several consecutive random layers is the same as the effect of one random layer: a random projection of the input. None of the work of disentangling the relevant factors of variation has been performed by these random layers and so the remaining trainable layers have the same job to do as was done by the full set of layers in the standard network. According to this hypothesis, the representational transformations originally performed by all 12 layers in the standard network must be somehow compressed into the remaining trained layers of the random networks. The hypothesis that these representational transformations will be evenly distributed across the remaining trainable layers is depicted in Figure~\ref{fig:hypothesis}. The performance of the random network would only be dependent on whether the structure and capacity of the remaining layers is sufficient to learn and represent the necessary transformations. Under this interpretation, a gradual degradation in performance as more layers are left untrained seems more likely and the sharp drop in performance observed in \citeA{Yosinski2014} is unexplained. To test this hypothesis, we calculated RV2 similarity matrices comparing each random network to the standard English network.

\begin{figure*}[h!]
\centering
\includegraphics[width=2\columnwidth]{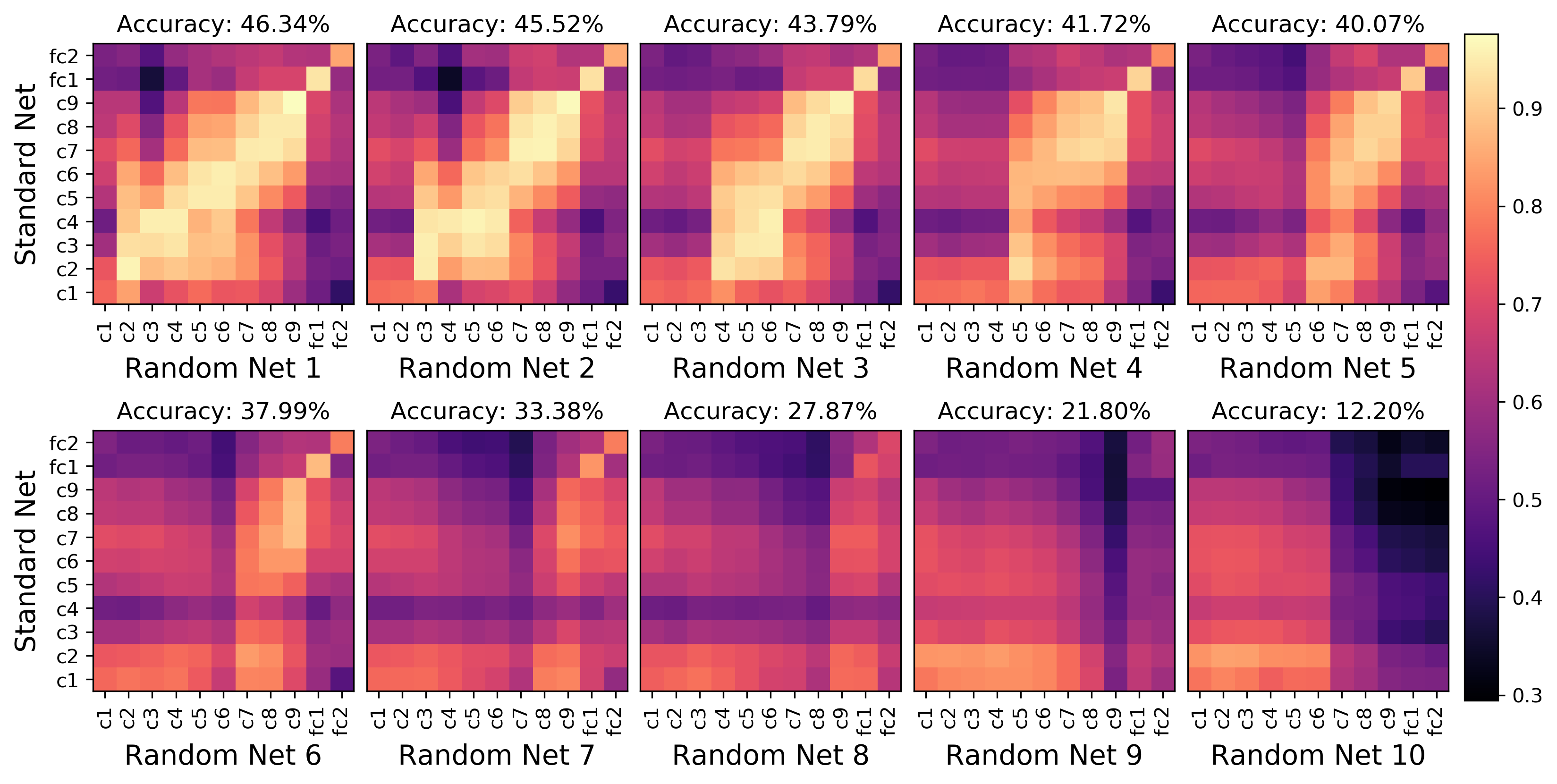}
\caption{\small The RV2 similarity between the baseline model (all layers trained) and networks of identical architecture with only layers above layer $n$ trained, $\forall n \in $ [c1, fc1]. } 
\label{fig:random-feats}
\end{figure*}

The comparisons between the standard model and the random networks are shown in Figure~\ref{fig:random-feats}. In the following, `random net $n$' refers to the network with random layers up to layer $n$; only layers above layer $n$ were trained. Layers are named c1, c2, ..., c9, fc1, fc2 to distinguish the convolutional and fully connected layers.
In contrast with our hypothesis, late layers remain most similar to their corresponding layer in the standard network, even as more early layers are left untrained. This pattern is especially clear in the similarity matrix for random net 4. The first trained layer of random net 4, layer c5, is diffusely similar to layers c2--c6 in the standard network, while the remaining layers show maximum similarity near the diagonal. 
When a network is mostly composed of random layers and only the fully connected layers are trained (e.g. random nets 9-10), the trained layers are not similar to any layer in the standard network. While these networks are still able to perform the task to some extent, they clearly do so in a way that does not mimic the standard network. 

\section{Discussion}
\citeA{Kornblith2019SimilarityRevisited} validated the CKA method by showing that it can identify corresponding layers in two networks trained from different random initializations. 
Our comparisons of freeze-trained networks, standard networks and untrained networks extend this validation by showing that a related similarity metric, RV2, applied to networks trained on speech, can recover expected and interpretable patterns of similarity.

%

Our random networks do not show an even distribution of the needed representational transformations across all trained layers. Instead, early trained layers compensate more for the reduced number of trained layers, such that the representations in late trained layers are less affected. 
%
This may reflect architectural constraints on representation. For example, fully connected layers may tend to be more similar to other fully connected layers than to convolutional layers and the fully connected layers may require a particular representation in the preceding convolutional layers. This top-down influence on representations in late layers may also be attributable to the targets serving as an anchor in the same way that the inputs anchor the representations in early layers. While there may be many computational solutions to the classification problem at hand, the form of the inputs and targets themselves are fixed, which may constrain the form of representations near the input and targets. 

\section{Acknowledgments}
Thanks to Nuance and Mitacs for their support and to Jo\~ao Felipe Santos and Guillaume Alain for helpful comments.

\bibliographystyle{apacite}

\setlength{\bibleftmargin}{.125in}
\setlength{\bibindent}{-\bibleftmargin}

\bibliography{references}

\end{document}